\documentclass{article}
\usepackage{ijcai21}
\usepackage{times}

\usepackage{soul}
\usepackage{url}
\usepackage[hidelinks]{hyperref}
\usepackage[utf8]{inputenc}
\usepackage[small]{caption}
\usepackage{graphicx}
\usepackage{amsmath}
\usepackage{amssymb}
\usepackage{multirow}
\usepackage{booktabs}
\usepackage{subfigure} 
\usepackage{pdfpages}

\title{Multi-Level Graph Encoding with Structural-Collaborative Relation Learning for Skeleton-Based Person Re-Identification}

\author{
Haocong Rao$^{1,2}$\and
Shihao Xu$^{1,2,3}$\and
Xiping Hu$^{1,2,3}$\footnotemark[1]\and
Jun Cheng$^{1,2}$\And
Bin Hu$^{4,3}$\footnotemark[1]\\
\affiliations
$^1$Shenzhen Institute of Advanced Technology, Chinese Academy of Sciences\\
$^2$The Chinese University of Hong Kong, Hong Kong\\
$^3$Lanzhou University\\
$^4$Beijing Institute of Technology\\
\emails
haocongrao@gmail.com,
xushh16@lzu.edu.cn,
huxp@lzu.edu.cn,
jun.cheng@siat.ac.cn,
bh@bit.edu.cn
}

\begin{document}

\maketitle
\footnotetext[1]{Corresponding authors}

\begin{abstract}
Skeleton-based person re-identification (Re-ID) is an emerging open topic providing great value for safety-critical applications. 
Existing methods typically extract hand-crafted features or model skeleton dynamics from the trajectory of body joints, while they rarely explore valuable relation information contained in body structure or motion.
To fully explore body relations, we construct graphs to model human skeletons from different levels, and for the first time propose a Multi-level Graph encoding approach with Structural-Collaborative Relation learning (MG-SCR) to encode discriminative graph features for person Re-ID.
Specifically, considering that structurally-connected body components are highly correlated in a skeleton, we first propose a \textit{multi-head structural relation layer} to learn different relations of neighbor body-component nodes in graphs, which helps aggregate key correlative features for effective node representations.
Second, inspired by the fact that body-component collaboration in walking usually carries recognizable patterns, we propose a \textit{cross-level collaborative relation layer} to infer collaboration between different level components, so as to capture more discriminative skeleton graph features.
Finally, to enhance graph dynamics encoding, we propose a novel \textit{self-supervised sparse sequential prediction} task for model pre-training, which facilitates encoding high-level graph semantics for person Re-ID. MG-SCR outperforms state-of-the-art skeleton-based methods, and it achieves superior performance to many multi-modal methods that utilize extra RGB or depth features. Our codes are available at \href{https://github.com/Kali-Hac/MG-SCR}{https://github.com/Kali-Hac/MG-SCR}.
\end{abstract}
\section{Introduction}
Person re-identification (Re-ID) aims to re-identify a specific person in different views or scenes, which assumes a crucial role in human tracking and authentication
\cite{vezzani2013people}.
Mainstream studies typically utilize RGB images \cite{zhang2019densely}, 
depth images \cite{karianakis2018reinforced}, 
or skeleton data \cite{liao2020model} for person Re-ID.
Compared with RGB-based and depth-based methods that rely on human appearances or silhouettes,
3D skeleton-based models represent human body and motion with 3D coordinates of key body joints, and they enjoy smaller data size and better robustness to factors such as scale and view \cite{han2017space}. Hence, exploiting 3D skeletons to perform person Re-ID has drawn surging attention \cite{rao2020self}. However, the way to model or extract discriminative features of human body from 3D skeleton data remains to be an open problem.

\begin{figure}
    \centering
    \scalebox{0.38}{
    \includegraphics{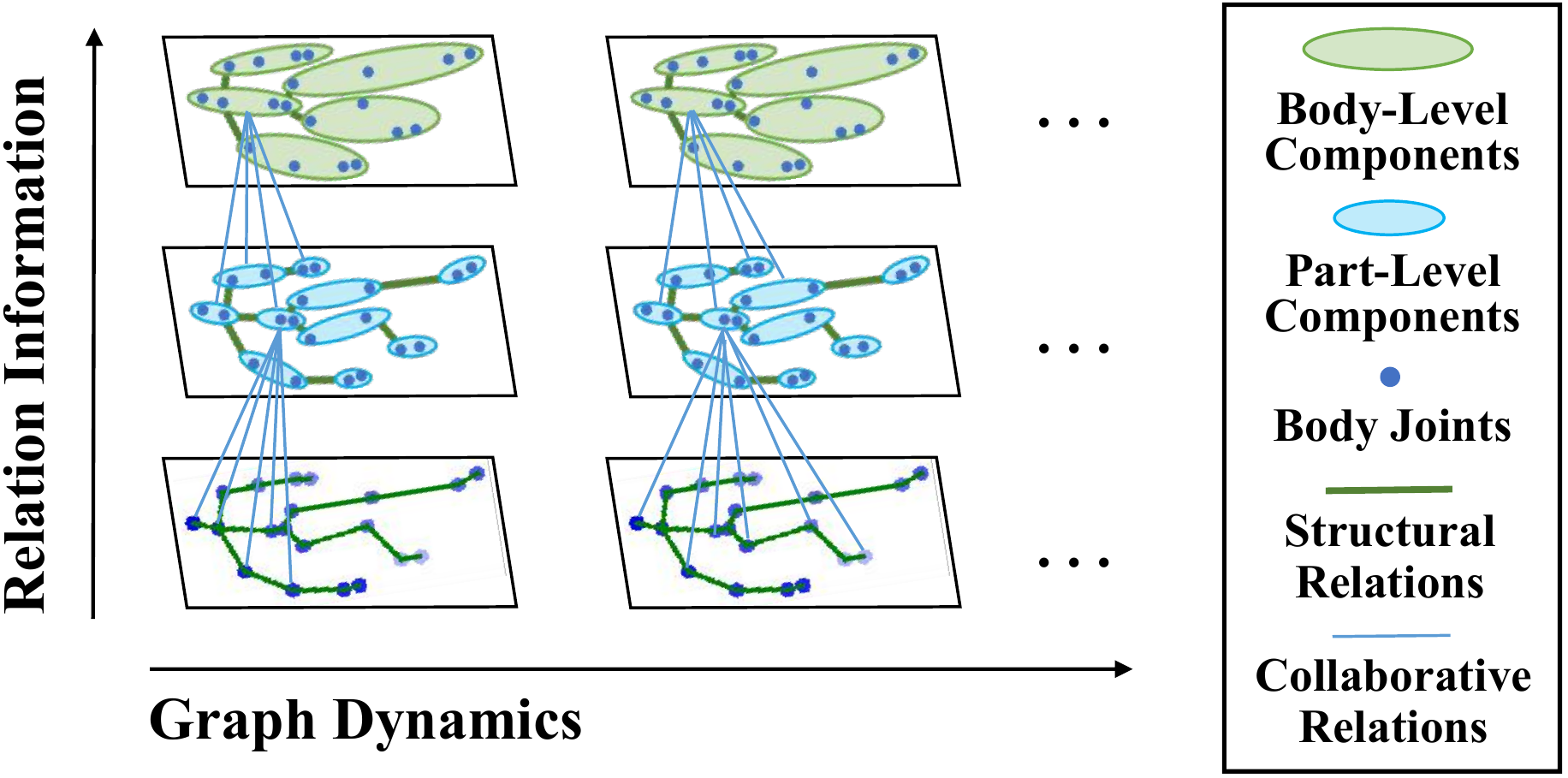}
    }
    \caption{Our approach constructs multi-level graphs for skeletons, and captures both skeleton graph dynamics and internal relation information (structural and collaborative relations) for person Re-ID.}
    \label{first_overview}
\end{figure}

Most existing methods manually design skeleton descriptors to depict certain discriminative attributes of body ($e.g.,$ gait and anthropometric attributes \cite{andersson2015person}) for person Re-ID. However, such hand-crafted methods heavily rely on domain knowledge like human anatomy, and typically lack the ability to mine latent features beyond human cognition. 
To alleviate this problem, recent works \cite{liao2020model,rao2020self} resort to deep neural networks (DNN) to perform representation learning of skeletons automatically.
These methods typically encode pairwise joint distances ($e.g.,$ limb lengths) or the trajectory of body-joint positions into a feature vector for modeling skeleton dynamics. However, they rarely explore latent relations between different body joints or components, thus ignoring valuable structural information of human body. 
Take people's walking for example, adjacent body joints such as ``knee'',``foot'' and collaborative limbs like ``arm'', ``leg'' usually possess different internal relations during movement, which could carry unique and recognizable walking patterns \cite{murray1964walking}.

To enable a full exploration of relations between different body components, this work for the first time constructs \textit{multi-level graphs} to represent each 3D skeleton at various levels, 
and proposes a Multi-level Graph encoding approach that learns both Structural and Collaborative Relations (MG-SCR) to encode discriminative body features for person Re-ID (illustrated in Fig. \ref{first_overview}). Specifically, considering that each body component is highly correlated with its physically-connected components and may possess different \textit{structural relations} ($e.g.,$ motion correlations), we first propose a \textbf{\textit{multi-head structural relation layer}} (MSRL) to capture multiple relations of one body-component node with respect to its neighbors in a graph, so as to focus on key correlative features and aggregate them to represent nodes. Second,
motivated by the fact that the cooperation of body components in walking could carry unique patterns ($i.e.$, gait) \cite{murray1964walking}, 
we propose a \textbf{\textit{cross-level collaborative relation layer}} (CCRL) to adaptively infer the \textit{degree of collaboration} between different level body components across graphs. By integrating graph features of adjacent levels via collaborative relations, CCRL encourages model to capture more structural semantics and discriminative skeleton features.
Third, to enhance graph dynamics encoding, we propose a novel \textit{self-supervised} pre-training task named \textbf{\textit{sparse sequential prediction}} (SSP) to exploit graph representations of \textit{unlabeled} skeleton subsequences for skeleton prediction, which facilitates capturing more high-level semantics ($e.g.,$ continuity of graphs) for person Re-ID. 
Finally, we fine-tune the SSP-pretrained model to predict ID labels for skeletons of a sequence, and leverage their average prediction to achieve effective person Re-ID.

In this paper, we make contributions as follows:
\begin{itemize}
\item We model 3D skeletons as multi-level graphs, and propose a novel multi-level graph encoding paradigm with structural-collaborative relation learning (MG-SCR) to encode discriminative graph features for person Re-ID.  

\item We propose multi-head structural relation layer (MSRL) to capture relations of neighbor body components, and devise cross-level collaborative relation layer (CCRL) to infer collaboration between different level components.

\item We propose a sparse sequential prediction (SSP) pre-training task to facilitate encoding graph dynamics and capturing high-level semantics for person Re-ID.

\end{itemize}

Extensive experiments on four datasets show that MG-SCR achieves state-of-the-art performance on skeleton-based person Re-ID. Besides, we provide a visualization to validate the ability of our model to infer internal relations between body components, and further demonstrate that MG-SCR is also effective with 3D skeleton data estimated from RGB videos.

\section{Related Works}
\paragraph{Skeleton-based Person Re-ID Methods.} Most existing works extract hand-crafted skeleton descriptors 
in terms of certain geometric, morphological or anthropometric attributes of human body: 
 \cite{barbosa2012re} calculates 7 Euclidean distances between the floor plane and joint or joint pairs to construct a distance matrix, which is learned by a quasi-exhaustive strategy to extract discriminative features. 
 \cite{munaro2014one} and \cite{pala2019enhanced} further extend them to 13 ($D^{13}$) and 16 skeleton descriptors ($D^{16}$) respectively, and leverage support vector machine (SVM), $k$-nearest neighbor (KNN) or Adaboost classifiers for Re-ID.
Since such solutions using 3D skeletons alone are hard to achieve satisfactory performance, they usually combine other modalities such as 3D point clouds \cite{munaro20143d} and 3D face descriptors \cite{pala2019enhanced} to improve Re-ID accuracy. Most recently, a few works exploit deep learning paradigms to learn gait representation from skeleton data: \cite{liao2020model} proposes PoseGait, which feeds 81 hand-crafted pose features of 3D skeletons into convolutional neural networks (CNN) for human recognition; \cite{rao2020self,rao2021a2} devise an attention-based gait encoding model with multi-layer long short-term memory (LSTM) \cite{hochreiter1997long} to learn gait features of skeletons in a self-supervised manner for person Re-ID. 

\paragraph{Depth-based and Multi-modal Person Re-ID Methods.} 
Depth-based methods typically extract human shapes, silhouettes or gait representations from depth images to perform person Re-ID. \cite{sivapalan2011gait} extends Gait Energy Image (GEI) \cite{chunli2010behavior} to 3D domain and proposes Gait Energy Volume (GEV) algorithm based on depth images to achieve gait-based human recognition. 
\cite{munaro2014one} extracts 3D point clouds from depth data and proposes a point cloud matching (PCM) method to discriminate different individuals via matching distances between multi-view point cloud sets. 
Multi-modal methods usually combine skeleton-based features with extra RGB or depth information such as depth shape features \cite{munaro20143d,wu2017robust,hasan2016long} to improve person Re-ID accuracy. In \cite{karianakis2018reinforced}, a split-rate RGB-depth transferred CNN-LSTM model with reinforced temporal attention (RTA) is proposed for person Re-ID task. 
 
 \begin{figure*}[ht]
    \centering
    \scalebox{0.565}{
    \includegraphics{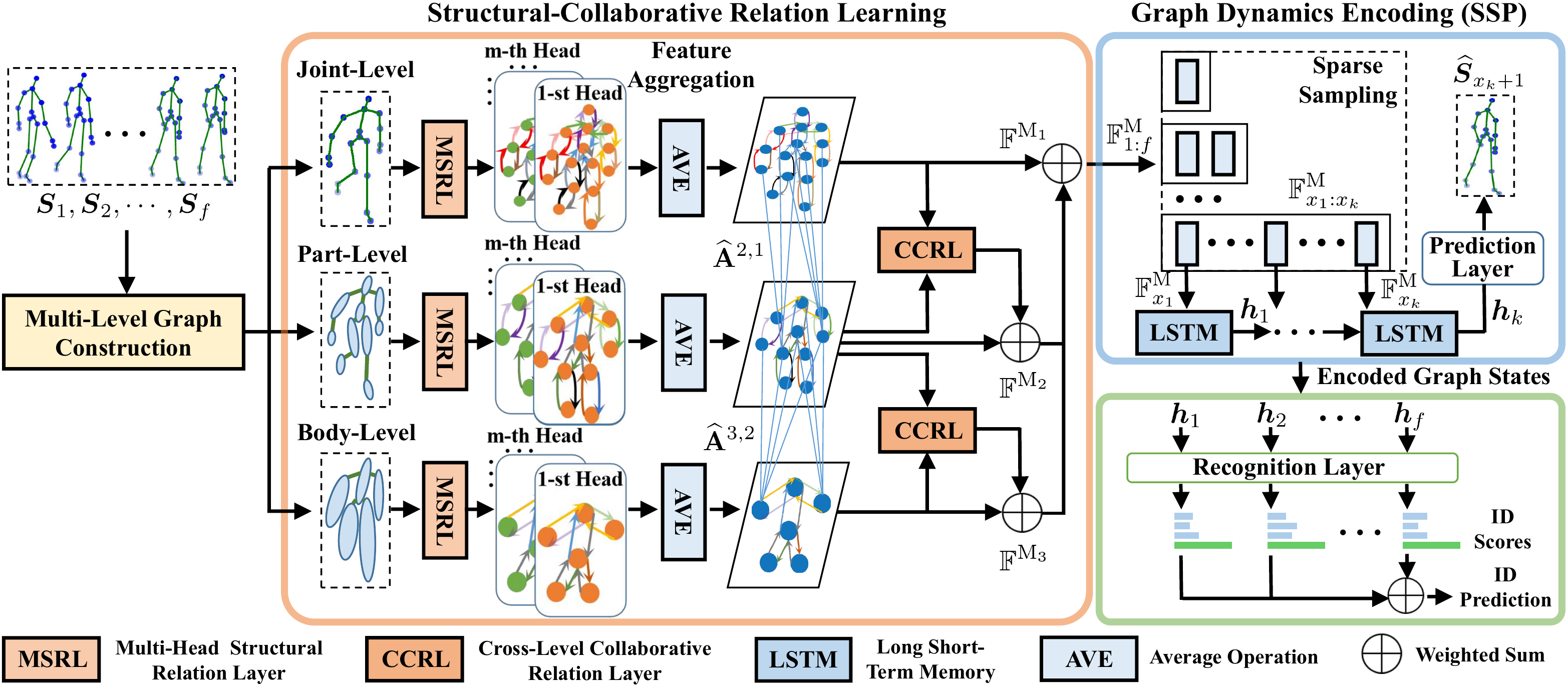}
    }
    \caption{Flow diagram of MG-SCR: (1) Each skeleton in a sequence $\boldsymbol{S}_{1},\cdots,\boldsymbol{S}_{f}$ is represented as joint-level, part-level and body-level graphs. (2) First, we employ multi-head structural relation layers (MSRL) to capture structural relations of neighbor nodes, and averagely aggregate features learned by multiple heads to represent nodes. (3) Then, cross-level collaborative relation layers (CCRL) infer collaboration between body components across adjacent graphs, namely $\mathbf{\widehat{A}}^{2,1}$ and $\mathbf{\widehat{A}}^{3,2}$, which are exploited to integrate graph features into multi-level graph representation $\mathbb{F}^{\text{M}}$. (4) Next, in SSP pre-training, we utilize LSTM to encode $\mathbb{F}^{\text{M}}_{x_{1}:x_{k}}$, which are multi-level graph representations of the sparsely sampled $k$-skeleton subsequence, into encoded graph states $\boldsymbol{h}$ to capture graph dynamics and predict next skeleton $\widehat{\boldsymbol{S}}_{x_{k}+1}$. (5) Finally, we feed encoded graph states $\boldsymbol{h}_{1},\cdots,\boldsymbol{h}_{f}$ of the input sequence into the recognition layer to fine-tune our model for person Re-ID.
    }
    \label{model_overview}
\end{figure*}
 
\section{The Proposed Approach}
Suppose that a 3D skeleton sequence $\boldsymbol{S}_{1:f}\!=\!(\boldsymbol{S}_1,\cdots,\boldsymbol{S}_{f})\in \mathbb{R}^{f \times J \times D}$, where $\boldsymbol{S}_{t}\in \mathbb{R}^{J \times D}$ is the $t^{th}$ skeleton with $J$ body joints and $D\!=\!3$ dimensions. The training set $\Phi=\left\{\boldsymbol{S}^{(i)}_{1:f}\right\}_{i=1}^{N}$ contains $N$ skeleton sequences collected from different persons and views. Each skeleton sequence $\boldsymbol{S}^{(i)}_{1:f}$ corresponds to an ID label $y_{i}$, where $y_{i}\in \{1, \cdots, C\}$ and $C$ is the number of different persons. Our goal is to predict the ID label of the input skeleton sequence: First, we construct multi-level graphs to represent each skeleton. Second, the proposed MG-SCR exploits multi-head structural relation layers and cross-level collaborative relation layers to capture different relations of graph nodes, and generates multi-level graph representations ($\mathbb{F}^{\text{M}}$) for skeletons in the sequence. Third, our model is pre-trained by a sparse sequential prediction task to encode dynamics of graph representations ($\mathbb{F}^{\text{M}}$) into encoded graph states ($\boldsymbol{h}$). Finally, we fine-tune the model with $\boldsymbol{h}$ to predict the sequence label $\hat{y}$ for person Re-ID. The overview of MG-SCR is given in Fig. \ref{model_overview}, and we present the details of each technical component below.

\subsection{Multi-Level Graph Construction}
\label{graph_construct}
Inspired by the fact that human motion can be decomposed into movements of several functional components ($e.g.,$ legs, arms) \cite{winter2009biomechanics}, we spatially group body joints, which are basic components, to be a higher level body-component node at the center of their positions.
As shown in Fig. \ref{model_overview}, we construct three levels of skeleton graphs, namely joint-level ($i.e.,$ body joints as nodes), part-level and body-level graphs for each skeleton $\boldsymbol{S}$, which can be represented as $\mathcal{G}^1, \mathcal{G}^2, \mathcal{G}^3$ respectively. 
Each graph $\mathcal{G}^{l}(\mathcal{V}^{l}, \mathcal{E}^{l})$  ($l\in\{1,2,3\}$) consists of nodes $\mathcal{V}^{l}=\{\boldsymbol{v}^{l}_{1}, \boldsymbol{v}^{l}_{2}, \cdots,\boldsymbol{v}^{l}_{n_l}\}$ ($\boldsymbol{v}^{l}_{i}\in\mathbb{R}^{D}$, $i\in\{1,\cdots,n_l\}$) and edges $\mathcal{E}^{l}=\{e^{l}_{i,j}\ | \boldsymbol{v}^{l}_{i}, \boldsymbol{v}^{l}_{j}\!\in\!\mathcal{V}^{l}\} $ ($e^{l}_{i,j}\in\mathbb{R}$). Here $\mathcal{V}^{l}$, $\mathcal{E}^{l}$ denote the set of nodes corresponding to different body components and set of their internal connection relations respectively, and $n_l$ denotes the number of nodes in $\mathcal{G}^{l}$. 
More formally, we define a graph's adjacency matrix as $\mathbf{A}^{l} \in \mathbb{R}^{n_l \times n_l}$ to represent the structural relations among $n_l$ nodes. Note that we compute the \textit{normalized} structural relations between node $i$ and its neighbors, $i.e.$, $\sum_{j \in \mathcal{N}_{i}}\mathbf{A}^{l}_{i,j}=1$, where $\mathcal{N}_{i}$ denotes neighbor nodes of node $i$ in $\mathcal{G}^{l}$. In the training stage, $\mathbf{A}^{l}$ is adaptively learned to capture flexible structural relations.

\subsection{Multi-Head Structural Relation Layer}
To learn an effective representation for each body-component node in skeleton graphs, it is desirable to focus on features of structurally-connected (neighbor) nodes, which enjoy higher correlations (referred as ``\textbf{\textit{structural relations}}'') than distant pairs. For instance, adjacent nodes usually have closer spatial positions and highly similar motion tendency. Therefore, we propose a multi-head structural relation layer (MSRL) to learn structural relations of neighbor nodes and aggregate the most correlative spatial features to represent each node. 

\paragraph{Structural Relation Head.} We first devise a basic structural relation head based on \textit{graph attention mechanism} \cite{velickovic2018graph}, which can focus on more correlated neighbor nodes by assigning larger attention weights, to capture the internal relation $e^{l}_{i, j}$ between adjacent nodes $i$ and $j$:
\begin{equation}
\label{eq_1}
e^{l}_{i,j}=\text {LeakyReLU}\!\left({\mathbf{W}^{l}_{r}}^{\mathsf{T}}\left[\mathbf{W}^{l}_{v}{\boldsymbol{v}}^{l}_{i} \| \mathbf{W}^{l}_{v}{\boldsymbol{v}}^{l}_{j}\right]\right)
\end{equation}
where $\mathbf{W}^{l}_{v}\in \mathbb{R}^{D_{1}\times D}$ denotes the weight matrix to map the $l^{th}$ level node features $\boldsymbol{v}^{l}_{i}\in \mathbb{R}^{D}$ into a higher level feature space $ \mathbb{R}^{D_1}$, $\mathbf{W}^{l}_{r}\in \mathbb{R}^{2D_{1}}$ is a learnable weight matrix to perform relation learning in the $l^{th}$ level graph,  $\|$ indicates concatenating features of two nodes, and LeakyReLU is a non-linear activation function. Then, to learn flexible structural relations to focus on more correlative nodes, we
normalize relations using the $\operatorname{softmax}$ function as following:
\begin{equation}
\label{eq_2}
\mathbf{A}^{l}_{i,j}= \operatorname{softmax}_{j}\left(e^{l}_{i,j} \right) = \frac{\exp \left(e^{l}_{i,j} \right)}{\sum_{k \in \mathcal{N}_{i}} \exp \left(e^{l}_{i,k} \right)}
\end{equation}
where $\mathcal{N}_{i}$ denotes directly-connected neighbor nodes (including $i$) of node $i$ in graph. We use structural relations $\mathbf{A}^{l}_{i,j}$ to aggregate features of most relevant nodes to represent node $i$:
\begin{equation}
\label{eq_aggr}
\boldsymbol{\overline{v}}^{l}_{i}=\sigma\left(\sum_{j \in \mathcal{N}_{i}} \mathbf{A}^{l}_{i,j} \mathbf{W}^{l}_{v} \boldsymbol{v}^{l}_{j}\right)
\end{equation}
Here $\sigma$ is a non-linear function and $\boldsymbol{\overline{v}}^{l}_{i}\in \mathbb{R}^{D_{1}}$ is feature representation of node $i$ computed by a structural relation head.

To sufficiently capture potential structural relations ($e.g.,$ position similarity, movement correlations) between each node and its neighbor nodes, we employ a \textit{multi-head structural relation layer} (MSRL), where each head independently executes the same computation of Eq. \ref{eq_aggr} to learn a different structural relation. We \textit{averagely aggregate} features learned by $m$ different heads as representation for node $i$ (see Fig. \ref{model_overview}):
\begin{equation}
\label{node_repre}
\boldsymbol{\widehat{v}}^{l}_{i}= \frac{1}{m} \sum^{m}_{s=1}\sigma\left( \sum_{j \in \mathcal{N}_{i}} (\mathbf{A}^{l}_{i,j})^{s} (\mathbf{W}^{l}_{v})^{s} \boldsymbol{v}^{l}_{j}\right)
\end{equation}where $\boldsymbol{\widehat{v}}^{l}_{i}\in\mathbb{R}^{D_{1}}$ denotes the multi-head feature representation of node $i$ in $\mathcal{G}^{l}$, $m$ is the number of structural relation heads, $(\mathbf{A}^{l}_{i,j})^{s}\in\mathbb{R}$ represents the structural relation between node $i$ and $j$ computed by the $s^{th}$ structural relation head, and $(\mathbf{W}^{l}_{v})^s$ denotes the corresponding weight matrix to perform feature mapping in the $s^{th}$ head. Here we use \textit{average} rather than concatenation operation to reduce feature dimension and allow for more structural relation heads. MSRL captures the structural relations of correlative neighbor nodes (see Eq. \ref{eq_1}, \ref{eq_2}) and integrates key spatial features into node representations of each graph (see Eq. \ref{eq_aggr}, \ref{node_repre}). However, it only considers the local relations within a graph and is insufficient to capture collaboration between different level components, which motivates us to propose a cross-level collaborative relation layer.

\subsection{Cross-Level Collaborative Relation Layer}
As our ultimate goal is to learn recognizable patterns of a skeleton sequence for person Re-ID, it is natural to consider the property of human walking---Gait, which could be represented by the dynamic cooperation among body joints or between different body components \cite{murray1964walking}. To exploit such nature to capture more discriminative walking patterns, when encoding a skeleton's multi-level graphs, we expect our model to infer the degree of collaboration (referred as ``\textbf{\textit{collaborative relations}}'') between a node and its spatially corresponding high-level body component or other potential components.
As shown in Fig. \ref{model_overview}, we propose a \textit{Cross-Level Collaborative Relation Layer} (CCRL) to compute collaborative relation matrix
 $\mathbf{\widehat{A}}^{l,l-1} \in \mathbb{R}^{n_l \times n_{l-1}}$ ($l \in \{2, 3\}$) between $l^{th}$ level nodes $\mathcal{V}^{l}$ and $(l-1)^{th}$ level nodes $\mathcal{V}^{l-1}$ as following: 
\begin{small}
\begin{equation}
\mathbf{\widehat{A}}^{l, l-1}_{i, j}\!=\!\operatorname{softmax}_{j}\left({\boldsymbol{\widehat{v}}^{l}_{i}}^{\top} \boldsymbol{\widehat{v}}^{l-1}_{j}\right)\!=\!\frac{\exp \left({\boldsymbol{\widehat{v}}^{l}_{i}}^{\top} \boldsymbol{\widehat{v}}^{l-1}_{j}\right)}{\sum^{n_{l-1}}_{k=1} \exp \left({\boldsymbol{\widehat{v}}^{l}_{i}}^{\top} \boldsymbol{\widehat{v}}^{l-1}_{k}\right)}
\end{equation}
\end{small}where $\mathbf{\widehat{A}}^{l, l-1}_{i, j}$ is the collaborative relation between node $i$ in $\mathcal{G}^l$ and node $j$ in $\mathcal{G}^{l-1}$. Here CCRL uses the inner product of multi-head node feature representations (see Eq. \ref{node_repre}), which retain key spatial information of nodes, to measure the degree of collaboration. Then, to adaptively focus on key correlative features in collaboration, we exploit the collaborative relations to update the $l^{th}$ level node representation $\widehat{\boldsymbol{v}}_{i}^{l}$ as below:
\begin{equation}
\label{eq_fuse}
\widehat{\boldsymbol{v}}_{i}^{l} \ \leftarrow \ \sum^{n_{l-1}}_{j=1}\mathbf{\widehat{A}}^{l, l-1}_{i, j}\ \mathbf{W}^{l-1}_{\text{c}} \  \boldsymbol{\widehat{v}}^{l-1}_{j} +\widehat{\boldsymbol{v}}_{i}^{l} 
\end{equation}
where $\mathbf{W}^{l-1}_{\text{c}}\in\mathbb{R}^{D_{1}\times D_{1}}$ is a learnable weight matrix to integrate features of collaborative node $\boldsymbol{\widehat{v}}^{l-1}_{j}$ into higher level one $\widehat{\boldsymbol{v}}_{i}^{l}$, and $n_{l-1}$ is the number of nodes in $(l-1)^{th}$ level graph. 

\paragraph{Multi-Level Graph Feature Fusion.} To combine all structural semantics of multiple graphs, we adopt a weighted sum of three level graph representations as the final multi-level graph representation. 
Inspired by \cite{li2020dynamic}, we 
\textit{broadcast} ($i.e.,$ replicate) each part-level or body-level node to match their corresponding body joints in joint-level graphs. Let the broadcast output graph features of three levels for \textit{a skeleton sequence} $\boldsymbol{S}_{1:f}$ be 
$\mathbb{F}^{\text{M}_{1}}, \mathbb{F}^{\text{M}_{2}}, \mathbb{F}^{\text{M}_{3}} \in \mathbb{R}^{f \times J \times D_1}$. We obtain the multi-level graph representation $\mathbb{F}^{\text{M}}$ as below:
\begin{equation}
\label{ml_rep}
\mathbb{F}^{\text{M}}=\mathbb{F}^{\text{M}_{1}}+\lambda\left(\mathbb{F}^{\text{M}_{2}}+\mathbb{F}^{\text{M}_{3}}\right)
\end{equation}
where $\lambda$ is the fusion coefficient to balance different levels, and $\mathbb{F}^{\text{M}}=(\mathbb{F}^{\text{M}}_{1}, \cdots, \mathbb{F}^{\text{M}}_{f})$,  $\mathbb{F}^{\text{M}}_{t} \in \mathbb{R}^{J \times D_{1}}$ denotes the multi-level graph representation of an input skeleton  $\boldsymbol{S}_{t}$ in $\boldsymbol{S}_{1:f}$. 

\subsection{Multi-Level Graph Dynamics Encoding}
Given multi-level graph representations 
($\mathbb{F}^{\text{M}}_{1}, \cdots, \mathbb{F}^{\text{M}}_{f}$)
of the skeleton sequence $\boldsymbol{S}_{1:f}$, we exploit an LSTM to integrate their temporal dynamics into effective representations: LSTM
encodes each graph representation $\mathbb{F}^{\text{M}}_{t}$ and the previous step's latent state $\boldsymbol{h}_{t-1}$ (if existed), which provides the temporal context information of graph representations, into the current latent state $\boldsymbol{h}_{t}$ ($t\in\{1, \cdots,f\}$) as follows:
\begin{equation}
\label{LSTM}
\boldsymbol{h}_{t}=\left\{\begin{array}{ll}
\phi\left(\mathbb{F}^{\text{M}}_{1}\right) & \text { if } \quad t=1 \\
\phi\left(\boldsymbol{h}_{t-1}, \mathbb{F}^{\text{M}}_{t}\right) & \text { if } \quad 1<t\leq f
\end{array}\right.
\end{equation}
where $\boldsymbol{h}_{t}\in \mathbb{R}^{D_2}$, $\phi(\cdot)$ denotes the LSTM encoder, which aims to capture long-term dynamics of graph representations.
$\boldsymbol{h}_{1}, \cdots, \boldsymbol{h}_{f}$ are \textit{encoded graph states} that contain crucial temporal encoding information of graph representations from time $1$ to $f$. Instead of directly utilizing the last encoded graph state $\boldsymbol{h}_{f}$ that compresses the temporal dynamics of a sequence \cite{weston2014memory}, we expect our model to mine latent high-level semantics ($e.g.,$ subsequence dynamics, continuity of graphs) and capture more discriminative features for person Re-ID. To this end, we propose a self-supervised sparse sequential prediction task to pre-train our model.

\begin{table*}[ht]
\centering

\scalebox{0.7}{
\setlength{\tabcolsep}{2.6mm}{
\begin{tabular}{cll|rrrrrrrrrr}
\specialrule{0.1em}{0.45pt}{0.45pt}
\multicolumn{1}{l}{}                                                                              &             & \multicolumn{1}{c|}{\textbf{}} & \multicolumn{2}{c}{\textbf{BIWI}}                                       & \multicolumn{2}{c}{\textbf{IAS-A}}                                      & \multicolumn{2}{c}{\textbf{IAS-B}}                                      & \multicolumn{2}{c}{\textbf{KGBD}}                                       & \multicolumn{2}{c}{\textbf{KS20}}                                       \\ \specialrule{0.1em}{0.45pt}{0.45pt}
\multicolumn{1}{l}{}                                                                              & \textbf{Id} & \textbf{Methods}               & \multicolumn{1}{c}{\textbf{Rank-1}} & \multicolumn{1}{c}{\textbf{nAUC}} & \multicolumn{1}{c}{\textbf{Rank-1}} & \multicolumn{1}{c}{\textbf{nAUC}} & \multicolumn{1}{c}{\textbf{Rank-1}} & \multicolumn{1}{c}{\textbf{nAUC}} & \multicolumn{1}{l}{\textbf{Rank-1}} & \multicolumn{1}{l}{\textbf{nAUC}} & \multicolumn{1}{l}{\textbf{Rank-1}} & \multicolumn{1}{l}{\textbf{nAUC}} \\ \specialrule{0.1em}{0.45pt}{0.45pt}
\multirow{4}{*}{\textbf{\begin{tabular}[c]{@{}c@{}}Depth-Based \\ Methods\end{tabular}}}          & 1           & Gait Energy Image  \shortcite{chunli2010behavior}            & 21.4                                & 73.2                              & 25.6                                & 72.1                              & 15.9                                & 66.0                              & —                                   & —                                 & —                                   & —                                 \\
                                                                                                  & 2  
                                          & 3D CNN + Average Pooling \shortcite{boureau2010theoretical}       & 27.8                                & 84.0                              & 33.4                                & 81.4                              & 39.1                                & 82.8                              & —                                   & —                                 & —                                   & —                                 \\ 
                                                                                                  & 3           
                                       & Gait Energy Volume  \shortcite{sivapalan2011gait}           & 25.7                                & 83.2                              & 20.4                                & 66.2                              & 13.7                                & 64.8                              & —                                   & —                                 & —                                   & —                                 \\                                                           & 4 
                            & 3D LSTM  \shortcite{haque2016recurrent}                      & 27.0                                & 83.3                              & 31.0                                & 77.6                              & 33.8                                & 78.0                              & —                                   & —                                 & —                                   & —                                 \\                  \specialrule{0.1em}{0.45pt}{0.45pt}
\multirow{6}{*}{\textbf{\begin{tabular}[c]{@{}c@{}}Multi-Modal \\  Methods\end{tabular}}}       & 5           & PCM + Skeleton  \shortcite{munaro20143d}                & 42.9                                & —                                 & 27.3                                & —                                 & 81.8                                & —                                 & —                                   & —                                 & —                                   & —                                 \\
                                                                                                  & 6           & Size-Shape descriptors + SVM \shortcite{hasan2016long}    & 20.5                                & 87.2                              & —                                   & —                                 & —                                   & —                                 & —                                   & —                                 & —                                   & —                                 \\
                                                                                                  & 7           & Size-Shape descriptors + LDA \shortcite{hasan2016long}   & 22.1                                & 88.5                              & —                                   & —                                 & —                                   & —                                 & —                                   & —                                 & —                                   & —                                 \\
                                                                                                  & 8           & DVCov + SKL \shortcite{wu2017robust}                    & 21.4                                & —                                 & 46.6                                & —                                 & 45.9                                & —                                 & —                                   & —                                 & —                                   & —                                 \\
                                                                                                  & 9           & ED + SKL \shortcite{wu2017robust}                      & 30.0                                & —                                 & 52.3                                & —                                 & 63.3                                & —                                 & —                                   & —                                 & —                                   & —                                 \\
                                                                                                  & 10          & CNN-LSTM with RTA \shortcite{karianakis2018reinforced}              & 50.0                                & —                                 & —                                   & —                                 & —                                   & —                                 & —                                   & —                                 & —                                   & —                                 \\ \specialrule{0.1em}{0.45pt}{0.45pt}
\multirow{7}{*}{\textbf{\begin{tabular}[c]{@{}c@{}}Skeleton-Based \\  Methods\end{tabular}}} & 11        
                                           & $D^{13}$ descriptors + KNN \shortcite{munaro2014one}          & 39.3                                & 64.3                              & 33.8                                & 63.6                              & 40.5                                & 71.1                              & 46.9                                & 90.0                              & 58.3                                & 78.0                              \\                                                          & 12          
                                           & Single-layer LSTM  \shortcite{haque2016recurrent}            & 15.8                                & 65.8                              & 20.0                                & 65.9                              & 19.1                                & 68.4                              & 39.8                                & 87.2                              & 80.9                                & 92.3                              \\
                                                                                                  & 13          & Multi-layer LSTM \shortcite{zheng2019relational}              & 36.1                                & 75.6                              & 34.4                                & 72.1                              & 30.9                                & 71.9                              & 46.2                                & 89.8                              & 81.6                                & 94.2                              \\
                                                                                                  & 14          & $D^{16}$ descriptors + Adaboost \shortcite{pala2019enhanced}     & 41.8                                & 74.1                              & 27.4                                & 65.5                              & 39.2                                & 78.2                              & 69.9                                & 90.6                              & 59.8                                & 78.8                              \\
                                                                                                  & 15          & PostGait \shortcite{liao2020model}                       & 33.3                                & 81.8                              & 41.4                                & 79.9                              & 37.1                                & 74.8                              & 90.6                                & 97.8                              & 70.5                                & 94.0                              \\
                                                                                                  & 16          & Attention Gait Encodings \shortcite{rao2020self}      & 59.1                                & 86.5                              & 56.1                                & 81.7                              & 58.2                                & 85.3                              & 87.7                                & 96.3                              & 86.5                                & 94.7                              \\
                                                                                                  & 17          & \textbf{MG-SCR (Ours)}                  & \textbf{61.6}                       & \textbf{91.9}                     & \textbf{56.5}                       & \textbf{87.0}                     & \textbf{65.9}                       & \textbf{93.1}                     & \textbf{96.3}                       & \textbf{99.9}                     & \textbf{87.3}                       & \textbf{95.5}                     \\ \specialrule{0.1em}{0.2pt}{0.2pt}
\end{tabular}
}
}
\caption{Comparison with existing skeleton-based methods (11-16). Depth-based methods (1-4) and multi-modal methods (5-10) are also included as a reference. Bold numbers refer to the best performers among skeleton-based methods. ``—'' indicates no published result.}
\label{results}
\end{table*}

\paragraph{Self-Supervised Sparse Sequential Prediction (SSP).}
 The aim of SSP is to enhance graph dynamics encoding and semantics learning by predicting future skeletons in a \textit{self-supervised} manner (note that SSP does NOT require any label to train):
First, we randomly sample a subsequence of length $k$ from the input sequence $\boldsymbol{S}_{1:f}$, and represent it as $\boldsymbol{S}_{x_1:x_k}\!=\!(\boldsymbol{S}_{x_1}, \cdots, \boldsymbol{S}_{x_k})$, where 
$x_{j}$ is $j^{th}$ sampled 
index and $j \leq x_{j}\leq f-1$.
Second, our model encodes $\boldsymbol{S}_{x_1:x_k}$ into graph representations $\mathbb{F}^{\text{M}}_{x_1:x_{k}}\!\!=\!\! (\mathbb{F}^{\text{M}}_{x_1},\cdots,\mathbb{F}^{\text{M}}_{x_k})$ (see Eq. \ref{eq_1}-\ref{ml_rep}), which are then fed into LSTM to generate encoded graph states $\boldsymbol{h}_{1}, \cdots, \boldsymbol{h}_{k}$ (see Eq. \ref{LSTM}). 
Last, we leverage 
$\boldsymbol{h}_{k}$ to predict the next skeleton with \textit{a prediction layer} $f_{pred}(\cdot)$ (see Fig. \ref{model_overview}):
\begin{equation}
\label{f_pred}
f_{pred}(\boldsymbol{h}_{k})=\widehat{\boldsymbol{S}}_{x_{k}+1}    
\end{equation}
where $\widehat{\boldsymbol{S}}_{x_{k}+1}\in \mathbb{R}^{J \times D}$ is the predicted $(x_{k}+1)^{th}$ skeleton, $f_{pred}(\cdot)$ is implemented by multi-layer perceptrons (MLPs).

 To exploit more potential samples for above prediction and semantics learning, we devise a \textit{sparse sampling scheme}: We randomly sample $f\!-\!1$ subsequences with lengths ($k$) from $1$ to $f\!-\!1$ respectively and make skeleton prediction for each subsequence $\boldsymbol{S}_{x_1:x_k}$ by Eq. \ref{f_pred}. In this way, we define the objective function $\mathcal{L}_{pred}$ for the self-supervision of SSP, which minimizes the mean square error (MSE) between the ground-truth skeleton and the predicted skeleton as following:
\begin{equation}
\label{pred_loss}
\mathcal{L}_{pred} = \frac{1}{N}\sum^{N}_{i=1}\sum^{f-1}_{k=1}\|\boldsymbol{S}^{(i)}_{l_k+1}-\widehat{\boldsymbol{S}}^{(i)}_{l_k+1}\|^{2}_{2} 
\end{equation}
where $l_{k}$ denotes the \textit{last} skeleton index in $k^{th}$ subsequence, 
$\boldsymbol{S}^{(i)}_{l_k+1},\widehat{\boldsymbol{S}}^{(i)}_{l_k+1}\in \mathbb{R}^{J\times D}$ 
are the $(l_{k}+1)^{th}$ ground-truth skeleton in $i^{th}$ input sequence and the predicted skeleton respectively.
$\|\cdot\|^{2}_{2}$ denotes square loss. To facilitate SSP learning, our optimization actually uses prediction loss of all skeletons: For a subsequence $\boldsymbol{S}_{x_1:x_k}$, we exploit its encoded graph states $\boldsymbol{h}_{1},\cdots, \boldsymbol{h}_{k}$ to predict $\widehat{\boldsymbol{S}}_{x_2},\cdots,\widehat{\boldsymbol{S}}_{x_{k}+1}$ respectively and compute the sum of all prediction loss.
 By learning to predict future positions and motion of skeletons dynamically ($i.e.,$ use various subsequences), SSP encourages integrating more crucial spatio-temporal features into encoded graph states to achieve better person Re-ID performance (see Sec. \ref{discussion}). 

\subsection{Recognition}
To perform person Re-ID, we feed encoded graph states
$\boldsymbol{h}_{1},\cdots,\boldsymbol{h}_{f}$ of the input sequence into \textit{a  recognition layer} $f_{re}(\cdot)$ built by MLPs to predict the sequence label. Specifically, we average the ID prediction of each encoded graph state  $f_{re}(\boldsymbol{h}_t)$ ($t \in \{1, \cdots, f\}$) in a sequence to be the final sequence-level ID prediction $\hat{y}$. 
We employ the cross-entropy loss to fine-tune the model with the recognition layer $f_{re}(\cdot)$:
\begin{equation}
\mathcal{L}_{re} =  -\frac{1}{N}\sum^{N}_{i=1}\sum^{C}_{j=1} y_{i,j}\operatorname{log} \hat{y}_{i,j}
 + \beta \|\Theta\|^{2}_{2}
\end{equation}
where $y_{i,j}$ is the ground-truth label ($y_{i,j}=1$ iff the $i^{th}$ skeleton sequence belongs to the $j^{th}$ class otherwise $0$), and $\hat{y}_{i,j}$ indicates the probability that the $i^{th}$ sequence is predicted as the $j^{th}$ class. Here $\Theta$ denotes the parameters of the model, and $ \beta \|\Theta\|^{2}_{2}$ is the  $L_{2}$ regularization with weight coefficient $\beta$.

\section{Experiments}
\subsection{Experimental Settings}
\paragraph{Datasets.} Our approach is evaluated on four public person Re-ID datasets that provide 3D skeleton data, namely \textit{IAS-Lab} \cite{munaro2014feature}, \textit{BIWI} \cite{munaro2014one}, \textit{KS20} \cite{nambiar2017context}, and \textit{KGBD} \cite{andersson2015person}, which contain skeleton data of 11, 50, 20, 164 different persons respectively. For IAS-Lab, BIWI and KGBD, we adopt the 
standard evaluation setup in \cite{rao2020self}. For KS20, since no training and testing splits are given, we randomly select one sequence from each viewpoint for testing and use the rest of skeleton sequences for training. 

To evaluate the effectiveness of our approach when 3D skeleton data are directly estimated from RGB videos rather than Kinect, we introduce a large-scale RGB video based dataset CASIA B \cite{yu2006framework}, which contains 124 individuals with 11 different views---$0^{\circ}$, $18^{\circ}$, $36^{\circ}$, $54^{\circ}$, $72^{\circ}$, $90^{\circ}$, $108^{\circ}$, $126^{\circ}$, $144^{\circ}$, $162^{\circ}$, $180^{\circ}$.
We follow \cite{liao2020model} and exploit pre-trained pose estimation models \cite{chen20173d,cao2019openpose} to extract 3D skeletons from RGB videos of CASIA B. We evaluate our approach on each view of CASIA B and use adjacent views for training. 

\paragraph{Implementation Details.} The number of body joints in the joint-level graph is $n_{1}=25$ in KS20, $n_{1}=14$ in CASIA B, and $n_{1}=20$ in other datasets. For part-level and body-level graphs, the numbers of nodes are $n_2=10$ and $n_3=5$ respectively. The sequence length $f$ on four skeleton-based datasets (BIWI, IAS-Lab, KGBD, KS20) is empirically set to $6$, which achieves best performance in average among different settings. For the largest dataset CASIA B with roughly estimated skeleton data from RGB frames, we set sequence length $f=20$ for training/testing.
The node feature dimension is $D_{1}=8$ and the number of heads in MSRL is $m=8$. We use $\lambda=0.3$ to fuse multi-level graph features. For graph dynamics encoding, we use a 2-layer LSTM with $D_{2} = 128$ hidden units per layer. 
We employ Adam optimizer with learning rate $0.0005$ for CASIA B and $0.005$ for other datasets. The batch size is $128$ for CASIA B and $256$ for other datasets. We set $L_{2}$ regularization coefficient to $0.0005$.

\paragraph{Evaluation Metrics.} 
Person Re-ID typically adopts a ``multi-shot'' manner that leverages predictions of multiple frames or a sequence representation to produce a sequence label. We compute Rank-1 accuracy and nAUC (area under the cumulative matching curve normalized by ranks \cite{gray2008viewpoint}) to evaluate multi-shot person Re-ID performance.

\subsection{Comparison with State-of-the-Art Methods}

 In Table \ref{results}, we compare our approach with state-of-the-art skeleton-based person Re-ID methods (Id = 11-16) on four datasets.  To provide a reference for the overall performance, we also include mainstream depth-based and multi-modal methods (Id = 1-10). The results are reported as below:

\paragraph{Comparison with Skeleton-based Methods.}
As presented in Table \ref{results}, our MG-SCR enjoys distinct advantages over existing skeleton-based methods in terms of Rank-1 accuracy and nAUC: First, compared with two most representative hand-crated methods (Id = 11, 14) that extract geometric skeleton descriptors $D^{13}$ and $D^{16}$, our model achieves a great improvement on Re-ID performance by $19.8\%$-$49.4\%$ Rank-1 accuracy and $9.3\%$-$27.6\%$ nAUC on all datasets.
Second, our approach significantly outperforms recent CNN-based (Id = 15) and LSTM-based models (Id = 12, 13, 16) by a large margin (up to $56.5\%$ Rank-1 accuracy and $26.1\%$ nAUC on different datasets).
In contrast to the PoseGait model (Id = 15) that requires manually extracting 81 pose and motion features for CNN learning, our model can automatically model spatial and temporal graph features from different levels, which facilitates capturing more discriminative features for person Re-ID. 
Besides, our MG-SCR also performs better than the latest Attention Gait Encodings (Id = 16) with a $7.7\%$-$8.6\%$ Rank-1 accuracy and $3.6\%$-$7.8\%$ nAUC gain on IAS-B and KGBD. On IAS-A and BIWI, despite both of them obtain a close Rank-1 accuracy, our approach can achieve an evidently higher nAUC by $5.4\%$ at least, which demonstrates the superior overall performance of our model on datasets that contain drastic shape and appearance changes (IAS-A and BIWI). 

\paragraph{Comparison with Depth-based Methods and Multi-modal Methods.} 
With 3D skeletons as the only input, the proposed MG-SCR outperforms classic depth-based methods (Id = 1-4) by more than $23.1\%$ Rank-1 accuracy and $5.6\%$ nAUC. Considering the fact that skeleton data are of much smaller data size than depth image data, our approach is both effective and efficient.
Compared with multi-modal methods (Id = 5-10) that exploit extra RGB or depth information, our MG-SCR is still the best performer in most cases. Notably, despite the multi-modal method (Id = 5) that uses both point cloud matching (PCM) and skeletons obtains the highest accuracy on IAS-B, it fails to yield satisfactory performance on datasets with a setting of frequent shape and appearance changes (IAS-A and BIWI). 
By contrast, our approach can achieve a better and more stable performance on each dataset, making it become a more promising person Re-ID solution. 


\begin{table}[t]
\centering

\scalebox{0.7}{
\setlength{\tabcolsep}{2.3mm}{
\begin{tabular}{ccccccccc}
\specialrule{0.1em}{0.45pt}{0.45pt}
 \multicolumn{2}{c}{\textbf{MSRL}}                                 & \textbf{CCRL} & \textbf{SSP} & $\boldsymbol{h}_{f}$ & \textbf{AP} & \textbf{Rank-1} & \textbf{nAUC} \\ \specialrule{0.1em}{0.45pt}{0.45pt}
 \textbf{Single-Level} & \multicolumn{1}{l}{\textbf{Multi-Level}} & \textbf{}    & \textbf{}    & \textbf{}     & \textbf{}        & \textbf{}       & \textbf{}     \\ \specialrule{0.1em}{0.45pt}{0.45pt}
 \checkmark                     &                                          &              &              &               & \checkmark                 & 56.8            & 89.1          \\
                     & \checkmark                                         &              &              &               & \checkmark                 & 57.3            & 89.9          \\
 \checkmark                      &                                          &              & \checkmark             & \checkmark              &                  & 56.9            & 89.6          \\
                       & \checkmark                                         &              & \checkmark             & \checkmark              &                  & 57.6            & 90.2          \\
 \checkmark                      &                                          &              & \checkmark             &               & \checkmark                 & 57.2            & 89.2          \\
                      & \checkmark                                         &              & \checkmark             & \textbf{}     & \checkmark                 & 59.3            & 91.0          \\
                       & \checkmark                                         & \checkmark             &              & \checkmark              &                  & 58.4            & 89.4          \\
                       & \checkmark                                        & \checkmark             &              &               & \checkmark                 & 59.1            & 90.6          \\
                       & \checkmark                                         & \checkmark             & \checkmark             & \checkmark              &                  & 59.7            & 91.0          \\
                       & \checkmark                                         & \checkmark             & \checkmark             &               & \checkmark                 & \textbf{61.6}   & \textbf{91.9} \\ \specialrule{0.1em}{0.2pt}{0.2pt}
\end{tabular}
}
}
\caption{Performance of our model with different components (MSRL, CCRL, SSP). ``Single-Level'' denotes using only joint-level graph. ``AP'' indicates exploiting average prediction of encoded graph states $\boldsymbol{h}_{1},\cdots,\boldsymbol{h}_{f}$ rather than final state $\boldsymbol{h}_{f}$ for person Re-ID. }
\label{components}
\end{table}

\section{Discussion}
\label{discussion}

\paragraph{Ablation Study.} 
 We perform ablation study to verify the effectiveness of each model component. As shown in Table \ref{components}, we draw the following conclusions:
\textbf{(a)} Exploiting multi-level graphs for person Re-ID can achieve better performance than merely using a joint-level graph by $0.5\%$-$2.1\%$ Rank-1 accuracy and $0.6\%$-$1.8\%$ nAUC, which justifies our claim that multi-level graphs are more effective skeleton representations of learning unique body features.  
\textbf{(b)} CCRL produces evident performance gain ($1.8\%$-$2.3\%$ Rank-1 accuracy and $0.7\%$-$0.9\%$ nAUC) when compared with utilizing MSRL solely. Such results demonstrate that CCRL can help capture more discriminative features via learning valuable collaborative relations for person Re-ID.
\textbf{(c)} Introducing SSP consistently improves the model performance by  $0.4\%$-$2.5\%$ Rank-1 accuracy and $0.1\%$-$1.6\%$ nAUC, 
which verifies the effectiveness of SSP on encoding more crucial graph dynamics to better perform person Re-ID task.
\textbf{(d)} Average prediction (AP) can boost Re-ID performance by up to $1.9\%$ Rank-1 accuracy
compared with directly using $\boldsymbol{h}_{f}$ for prediction. 
By reducing influence of noisy skeleton representations that give wrong predictions, AP encourages better sequence-level predictions. Other datasets report similar results. 

\paragraph{Evaluation with Model-estimated Skeletons.} 
To further evaluate our approach with model-estimated 3D skeletons instead of Kinect-based skeleton data, we exploit pre-trained pose estimation models \cite{cao2019openpose,chen20173d} to extract 3D skeletons from RGB videos of CASIA B, and compare the performance of MG-SCR with the state-of-the-art method PoseGait \cite{liao2020model}. As shown in Table \ref{CVE_comparison}, our approach outperforms PoseGait  with a large margin by $7.8\%$-$61.5\%$ Rank-1 accuracy on all views of CASIA B.
It is worth noting that MG-SCR can obtain more stable performance than PoseGait on 8 different continuous views from $18^\circ$ to $144^\circ$, which suggests that our approach possesses higher robustness to view-point variation. On two most challenging views ($0^\circ$ and $180^\circ$), our approach can also achieve superior performance to PoseGait by $9.3\%$-$21.2\%$ Rank-1 accuracy.
These results verify the effectiveness of MG-SCR on skeleton data estimated from RGB videos, and also show its great potential to be applied to large-scale RGB-based datasets under general settings ($e.g.,$ varying views).

\begin{figure*}[t]
    \centering
    \quad  \subfigure[CR Visualization (BIWI)]{    \scalebox{0.3}{\label{CCG_visual_1}\includegraphics[]{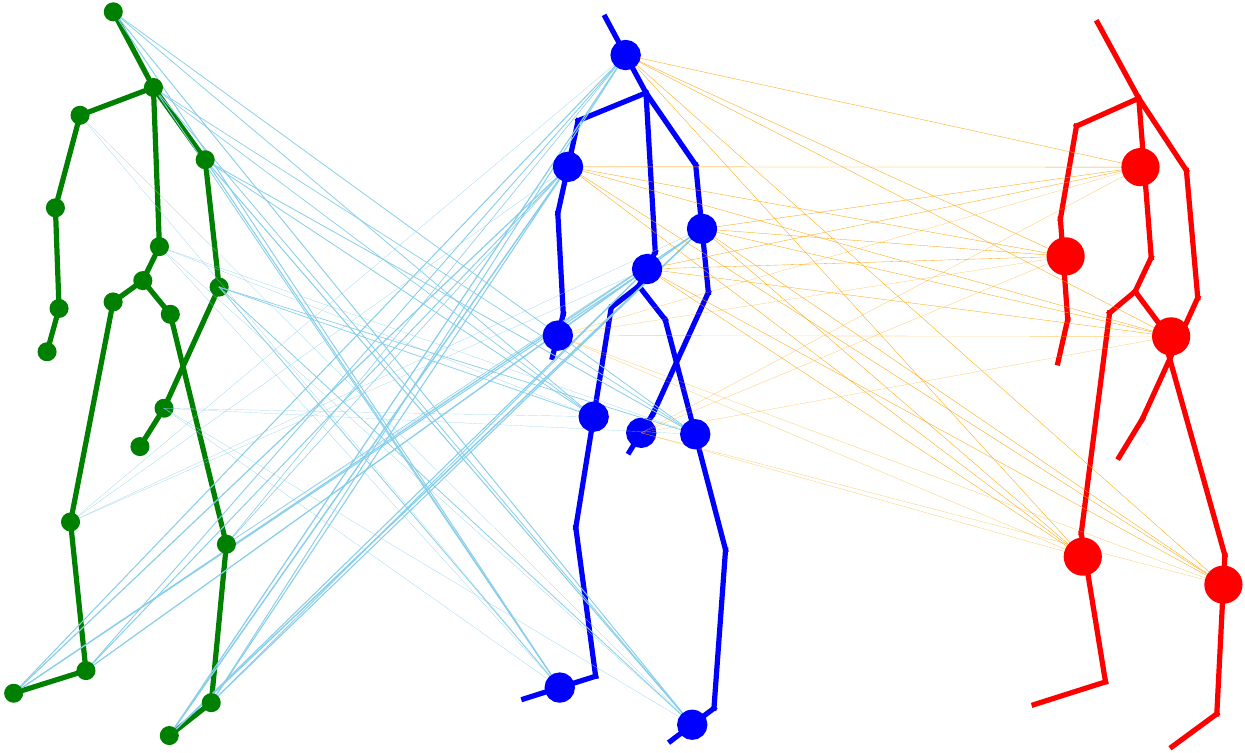}}
     }
     \quad \quad \quad \quad \quad \quad \quad \quad \quad \quad \quad \quad \quad \ \
    \subfigure[CR Visualization (KS20)]{    \scalebox{0.3}{\label{CCG_visual_2}\includegraphics[]{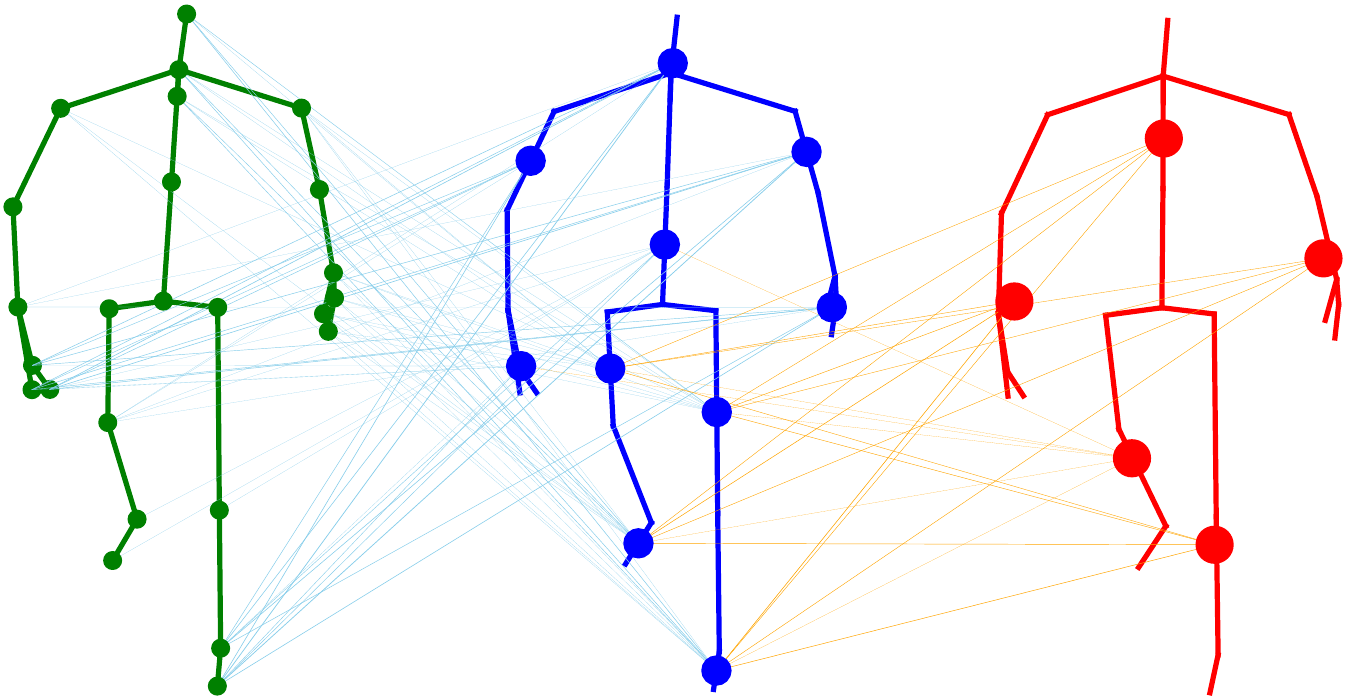}}
     }
    \subfigure[CR Matrix (BIWI)]{\scalebox{0.47}{\label{CCG_matrix_1}\includegraphics[]{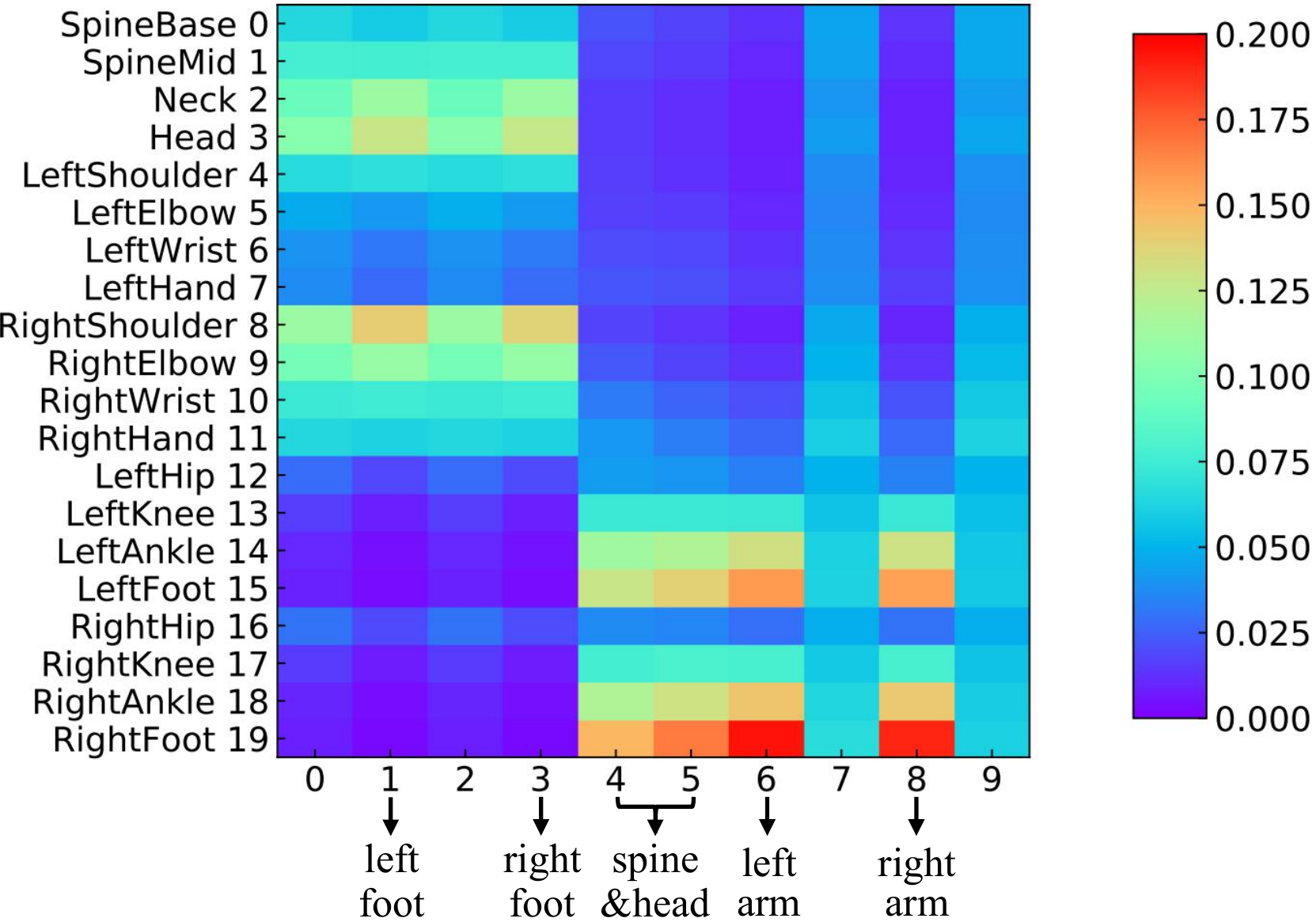}}
    }
   \quad \quad \quad 
    \subfigure[CR Matrix (KS20)]{\scalebox{0.47}{\label{CCG_matrix_2}\includegraphics[]{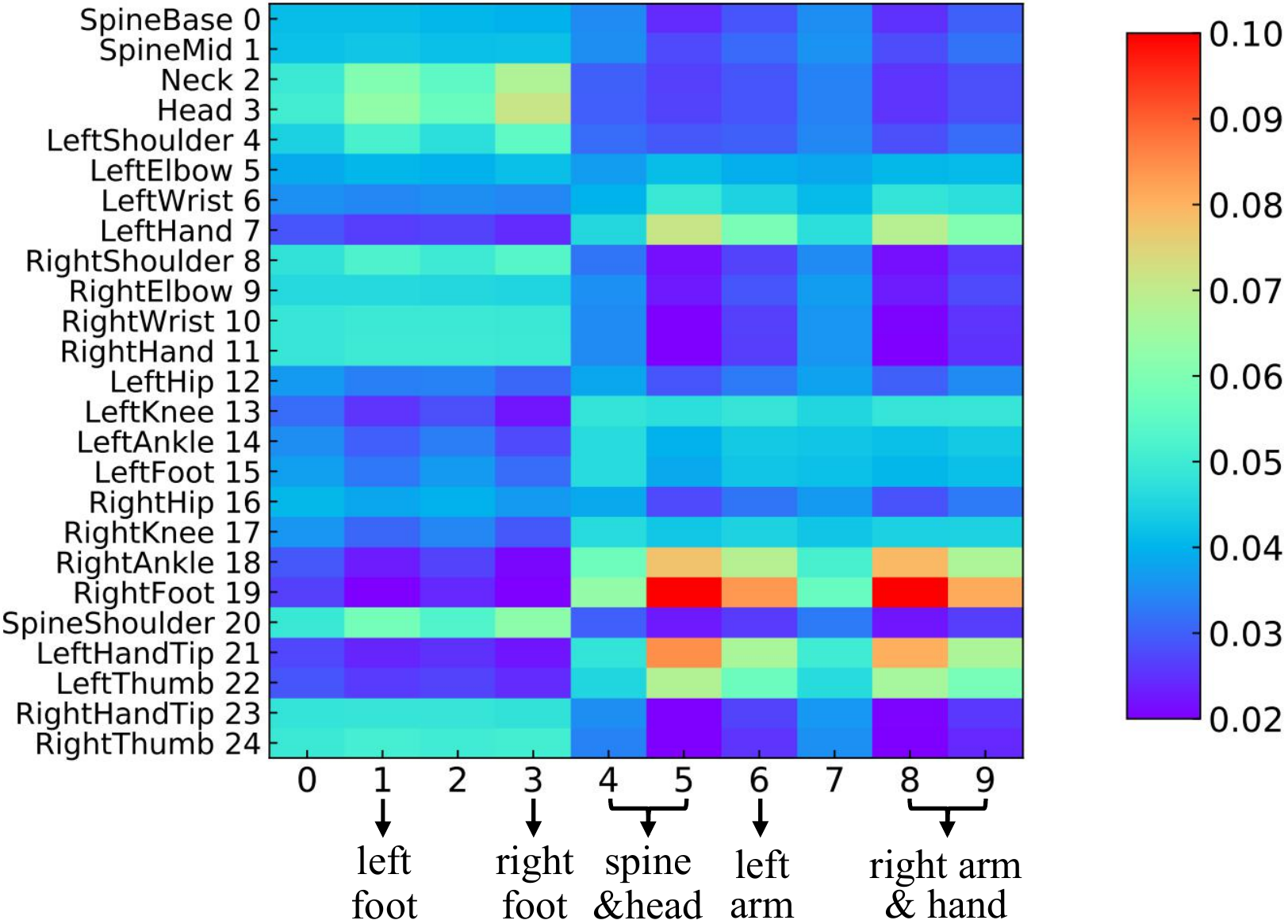}}  
    } 
    \caption{(a)-(b): Visualization of collaborative relations (CR) between different level body components for sample skeletons in BIWI and KS20 datasets. (c)-(d): CR matrices ($\widehat{\mathbf{A}}^{2, 1}$) between part-level ($\mathcal{G}^2$) and joint-level graphs ($\mathcal{G}^1$) for (a) and (b) respectively. Note that abscissa and ordinate denote indices of nodes and corresponding body components in $\mathcal{G}^2$ and $\mathcal{G}^1$. }
    \label{visual}
\end{figure*}

\begin{table}[t]
\centering
\scalebox{0.735}{
\setlength{\tabcolsep}{1mm}{
\begin{tabular}{lccccccccccc}
\specialrule{0.1em}{0.45pt}{0.45pt}
\multicolumn{1}{l}{\textbf{Methods}} & $\boldsymbol{0^{\circ}}$    & $\boldsymbol{18^{\circ}}$   & $\boldsymbol{36^{\circ}}$   & $\boldsymbol{54^{\circ}}$   & $\boldsymbol{72^{\circ}}$   & $\boldsymbol{90^{\circ}}$   & $\boldsymbol{108^{\circ}}$  & $\boldsymbol{126^{\circ}}$  & $\boldsymbol{144^{\circ}}$  & $\boldsymbol{162^{\circ}}$  & $\boldsymbol{180^{\circ}}$  \\ \specialrule{0.1em}{0.45pt}{0.45pt}
PoseGait                           & 10.7          & 37.4          & 52.5          & 28.3          & 24.3          & 18.9          & 23.5          & 17.2          & 23.6          & 18.8          & 4.3           \\
\textbf{Ours}                        & \textbf{20.0} & \textbf{63.1} & \textbf{60.3} & \textbf{52.0} & \textbf{54.0} & \textbf{80.4} & \textbf{75.1} & \textbf{74.3} & \textbf{65.6} & \textbf{39.1} & \textbf{25.5} \\ \specialrule{0.1em}{0.2pt}{0.2pt}
\end{tabular}
}
}
\caption{Rank-1 accuracy on different views of CASIA B.}
\label{CVE_comparison}
\end{table}

\paragraph{Analysis of Collaborative Relations.}As shown in Fig. \ref{visual}, we visualize node positions and collaborative relations of multi-level graphs (note that we draw relations with $\widehat{\mathbf{A}}_{i, j}^{l, l-1}$ larger than $0.065$, $0.045$ in Fig. \ref{CCG_visual_1} and Fig. \ref{CCG_visual_2}), and we obtain observations as follows: \textbf{(a)} There are distinct relations between different moving body components such as arms and legs, and CCRL learns stronger relations for the significantly collaborative components (see Fig. \ref{CCG_matrix_1}, \ref{CCG_matrix_2}), which verifies its ability to infer the dynamic cooperation of body components.
\textbf{(b)} Low level collaborative relations between $\mathcal{G}^1$ and $\mathcal{G}^2$ can capture global collaboration between different joints and body components, while the high level ones between $\mathcal{G}^2$ and $\mathcal{G}^3$ focus on certain ($i.e.,$ upper or lower) limbs. 


\section{Conclusion}
In this paper, we construct skeleton graphs at various levels and propose a novel multi-level graph encoding paradigm based on structural-collaborative relation learning (MG-SCR) to encode discriminative graph features for person Re-ID. 
We propose the multi-head structural relation layer to capture relations of neighbor body-component nodes and 
aggregate key features to effectively represent nodes. 
To capture more discriminative walking patterns, we devise the cross-level collaborative layer to explore dynamic collaboration between different level body components. A sparse sequential prediction pre-training task is proposed to enhance graph dynamics encoding for person Re-ID.
MG-SCR outperforms state-of-the-art methods on skeleton-based person Re-ID, and it obtains superior performance to many multi-modal methods.

\section*{Acknowledgements}
This work was supported in part by the National Key Research and Development Program of China (Grant No. 2019YFA0706200), and in part by the National Natural Science Foundation of China (Grant No. 61632014, 61627808).

\bibliographystyle{named}
\bibliography{ijcai21}
\newpage
\setcounter{section}{0}
\renewcommand\thesection{\Alph{section}}
\section{Appendix}

\begin{figure}[ht]
    \centering
    \scalebox{0.4}{
    \includegraphics{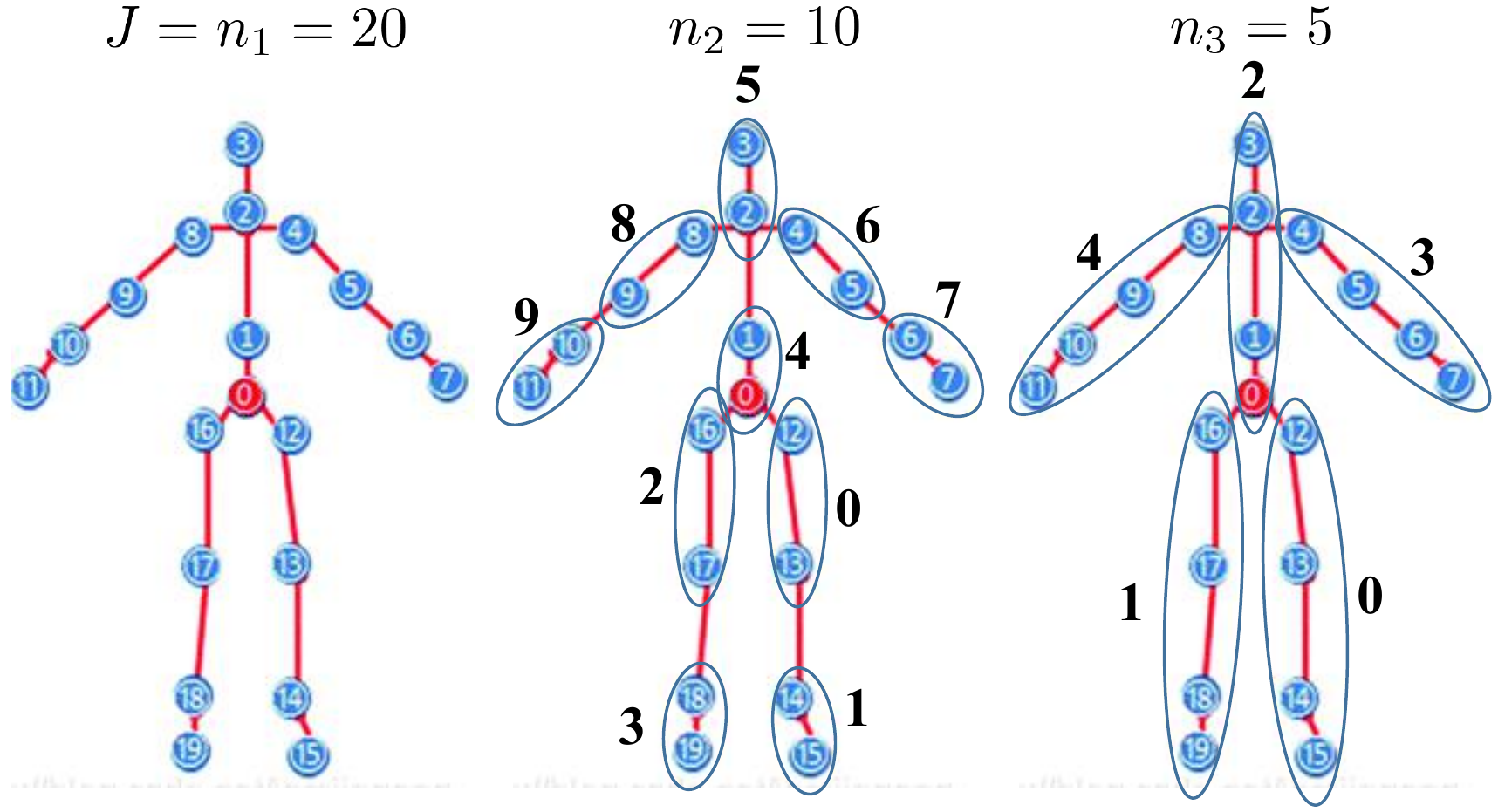}
    }
    \caption{Node indices (top) for joint-level ($n_1=20$), part-level ($n_2=10$), and body-level ($n_3=5$) graphs (bottom) on BIWI, IAS-Lab, and KGBD datasets. We spatially group body joints to be a more abstract body component in their central position ($i.e.,$ the average position of all body joints in a group).} 
    \label{node_indices}
\end{figure}

\begin{figure}[ht]
    \centering
    \scalebox{0.4}{
    \includegraphics{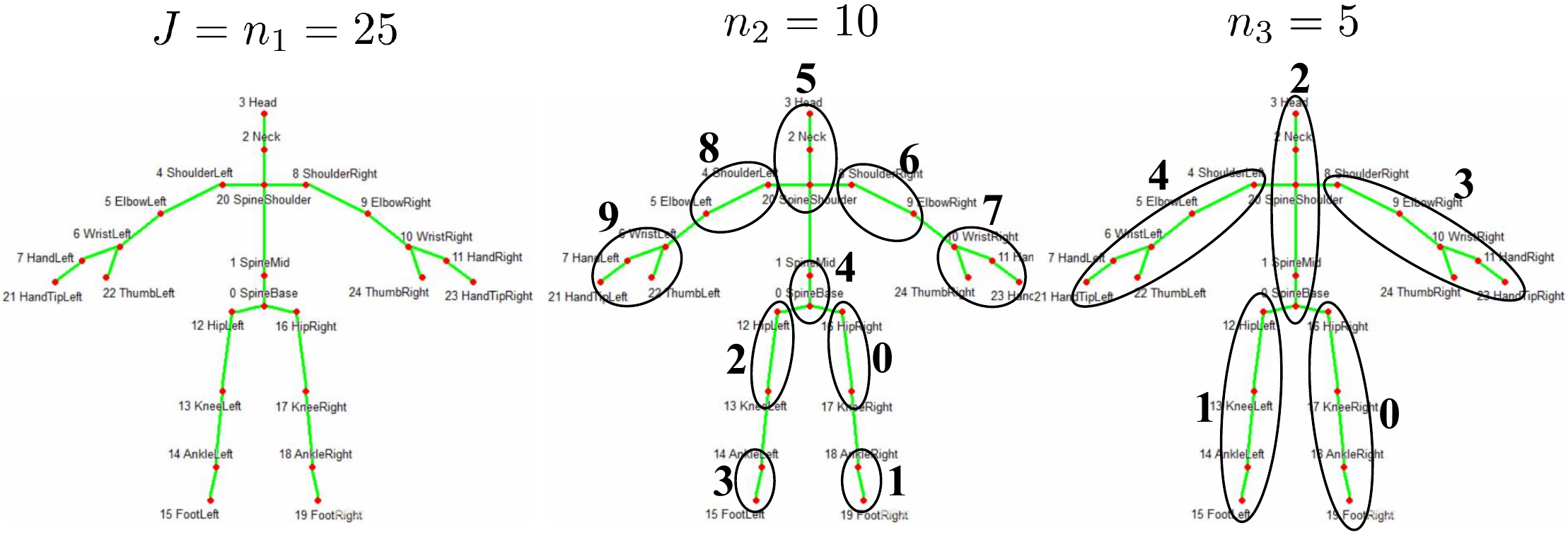}
    }
    \caption{Node indices (top) for joint-level ($n_1=25$), part-level ($n_2=10$), and body-level ($n_3=5$) graphs (bottom) on KS20 dataset. We spatially group body joints to be a more abstract body component in their central position ($i.e.,$ the average position of all body joints in a group).} 
    \label{node_indices_2}
\end{figure}

\begin{figure}[ht]
    \centering
    \scalebox{0.25}{
    \includegraphics{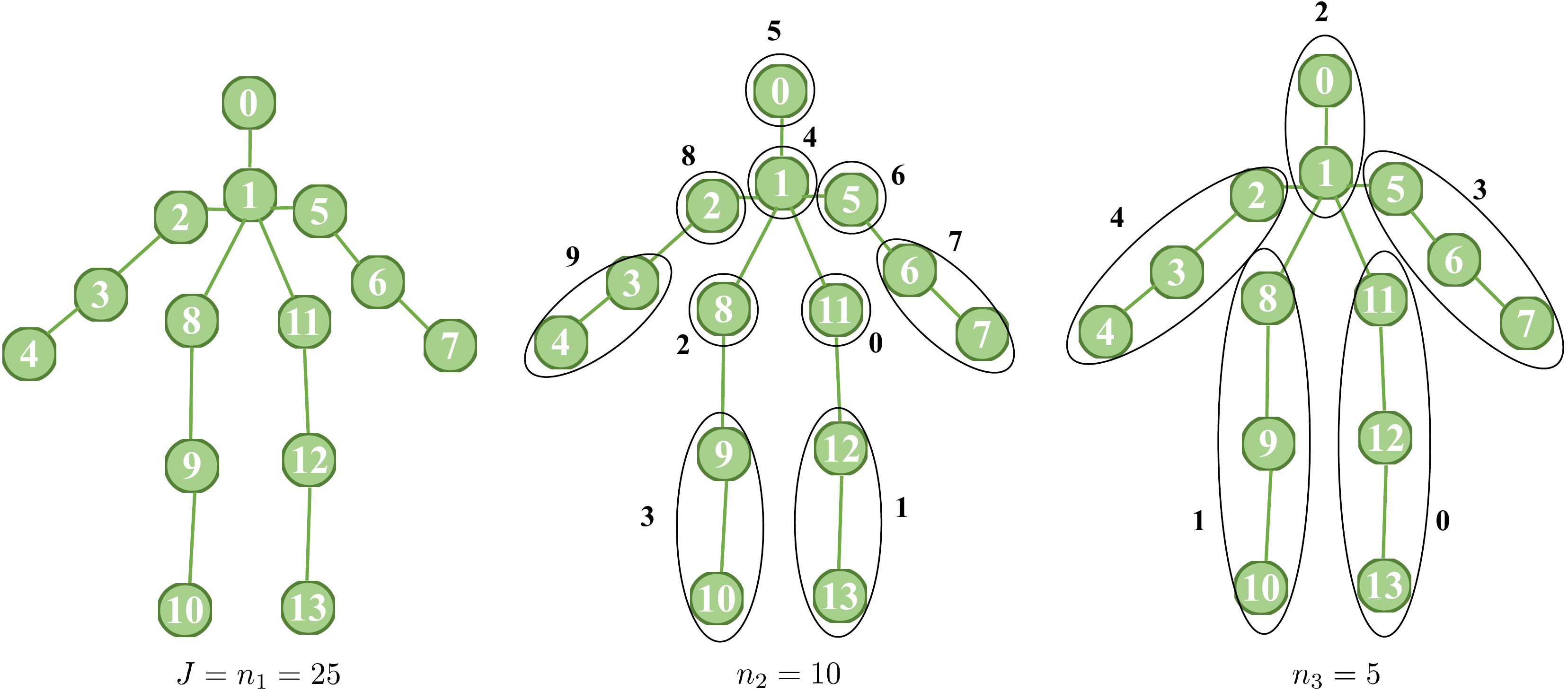}
    }
     \caption{Node indices for joint-scale ($n_1=14$), part-scale ($n_2=10$), and body-scale ($n_3=5$) graphs on CASIA B dataset. Note: All 3D skeletons are estimated from RGB videos (see Sec. \ref{3D_estimate}).} 
    \label{node_indices_3}
\end{figure}

\subsection{3D Skeleton Estimation Setup for CASIA B}
\label{3D_estimate}
For the RGB-based dataset CASIA B, we first extract eighteen 2D joints from each person in videos by \textit{OpenPose} model \cite{cao2019openpose}. Then, we follow the same configuration of estimation in \cite{liao2020model} and average the positions of ``Nose'', ``Reye'', ``LEye'', ``Rear'' and ``Lear'' as the position of ``Head'' to construct fourteen 2D joints (see Fig. \ref{node_indices_3}), which are fed into pose estimation method \cite{chen20173d} to estimate corresponding 3D body joints. Thus, $J$ is 14 ($i.e.,$ $n_{1}=14$) for CASIA B, and we normalize all joints in each skeleton by subtracting the neck joint. 

\subsection{Datatset Splitting} We adopt the standard evaluation setup in \cite{haque2016recurrent,rao2020self}: For IAS-Lab, we use the full training set and IAS-A/B test splits; For BIWI, we use the \textit{Walking} testing set and its corresponding training set that contains dynamic skeleton data; For KGBD, since no training and testing split are given, we randomly leave one skeleton video of each person for testing and use the rest of videos for training.
For KS20, we randomly select one skeleton sequence from each viewpoint for testing and use the rest of sequences for training. For CASIA B, we evaluate our approach on each view and use the adjacent views for training.

\subsection{Data Preprocssing}
 To avoid ineffective skeleton recording, We discard the first and last 10 skeleton frames of each original skeleton sequence. We normalize all skeleton sequences by subtracting the spine joint position from each joint of the same skeleton [Zhao \textit{et al.}, 2019].
 Then, we spilt all normalized skeleton sequences in the dataset into multiple shorter skeleton sequences ($i.e.$, $\boldsymbol{S}^{(i)}_{1:f}$) with length $f$ by a step of $\frac{f}{2}$, which aims
to obtain as many 3D skeleton sequences as possible to train our model. Unless explicitly specified, the skeleton sequence $\boldsymbol{S}^{(i)}_{1:f}$ in our paper refers to those split and normalized sequences used in learning, rather than those original skeleton sequences provided by datasets. 

\subsection{Experimental Details}
 As shown in Fig. \ref{node_indices}, Fig. \ref{node_indices_2} and Fig. \ref{node_indices_3}, The number of body joints in the joint-level graph is $n_{1}=25$ in KS20, $n_{1}=14$ in CASIA B, and $n_{1}=20$ in other datasets. For part-level and body-level graphs, the numbers of nodes are $n_2=10$ and $n_3=5$ respectively. For IAS-Lab, KS20 and KGBD datasets, the sequence length $f$ is empirically set to $6$, which achieves best performance in average among different settings. As to CASIA B, it is a large-scale dataset with roughly estimated skeleton data from RGB frames, which is intrinsically different from previous datasets. Our experiments show that a small $f$ like 6 performs poorly not only with our method, but also with other state-of-the-art methods like PoseGait \cite{liao2020model}. Therefore, we choose a different $f$ value ($f= 20$) for this dataset.
The node feature dimension is $D_{1}=8$ and the number of heads in MSRL is $m=8$. We apply the LeakyReLU nonlinearity with negative input slope $\alpha=0.2$ \cite{velickovic2018graph}, and employ ELU [Clevert \textit{et al.,} 2016] as the non-linear activation function $\sigma(\cdot)$ in structural relation learning.
We use $\lambda=0.3$ to fuse multi-level graph features. For graph dynamics encoding, we use a 2-layer LSTM with $D_{2} = 128$ hidden units per layer. 
We employ Adam optimizer with learning rate $0.0005$ for CASIA B and $0.005$ for other datasets. The batch size is $128$ for CASIA B and $256$ for other datasets. We set $L_{2}$ regularization coefficient to $0.0005$.
 Interested readers can access our source code at \href{https://github.com/Kali-Hac/MG-SCR}{https://github.com/Kali-Hac/MG-SCR} to get more details.

Our model is implemented with Tensorflow [Abadi \textit{et al.}, 2016] v1.9 and trained on the NVIDIA TITAN V GPU and Intel(R) Xeon(R) E5-2690 CPU.
We pre-train our model with SSP for 300 epochs on KS20, 200 epochs on CASIA B and 150 epochs for other datasets, then fine-tune our model with a recognition layer for person Re-ID. To avoid over-fitting and achieve better generalization performance, we adopt Early Stopping [Prechelt \textit{et al.}, 1998] with a patience of 100 epochs ($i.e.$, stop the training of model after no improvement in 100 continuous epochs). In Fig. \ref{node_indices}, we provide the indices of nodes in skeleton graphs at each level. Note that there are 25 body joints in a skeleton in KS20 dataset while other datasets have 20 joints. These body joints and their corresponding body components share the same indices:  (0) SpineBase, (1) SpineMid, (2) Neck, (3) Head, (4) LeftShoulder, (5) LeftElbow, (6) LeftWrist, (7) LeftHand, (8) RightShoulder, (9) RightElbow, (10) RightWrist, (11) RightHand, (12) LeftHip, (13) LeftKnee, (14) LeftAnkle, (15) LeftFoot, (16) RightHip, (17) RightKnee, (18) RightAnkle, (19) RightFoot, (20) SpineShoulder, (21) LeftHandTip, (22) LeftThumb, (23) RightHandTip, (24) RightThumb (see nodes 0-19 in the joint-level graph of Fig. \ref{node_indices} and nodes 0-24 in the joint-level graph of Fig. \ref{node_indices_2}). As to the estimated skeleton data from RGB videos of CASIA B, there are only 14 body joints, namely (0) Head, (1) Neck, (2) RShoulder, (3) RElbow, (4) RWrist, (5) LShoulder, (6) LElbow, (7) LWrist, (8) RHip, (9) RKnee, (10) Rankle, (11) LHip, (12) LKnee, (13) LAnkle (see nodes 0-13 in the joint-scale graph of Fig. \ref{node_indices_3}).

\end{document}